\documentclass[conference]{IEEEtran}
\IEEEoverridecommandlockouts
\usepackage{cite}
\usepackage{amsmath,amssymb,amsfonts}
\usepackage{amsthm}
\usepackage{algorithmic}
\usepackage{graphicx}
\usepackage{textcomp}
\usepackage{xcolor}

\usepackage{multirow}
\usepackage{booktabs}
\usepackage[algo2e,ruled,linesnumbered]{algorithm2e}

\def\BibTeX{{\rm B\kern-.05em{\sc i\kern-.025em b}\kern-.08em
    T\kern-.1667em\lower.7ex\hbox{E}\kern-.125emX}}
\begin{document}

\title{A Few-Shot LLM Framework for Extreme Day Classification in Electricity Markets  \\

\thanks{The authors would like to acknowledge supports from Columbia Data Science Institute, Red River Clean Energy, King Abdulaziz City for Science and Technology (KACST), and National Science Foundation under award ECCS-2239046.}
}

\author{\IEEEauthorblockN{Saud Alghumayjan, Ming Yi, Bolun Xu}
\IEEEauthorblockA{\textit{Dept. of Earth and Environmental Engineering} \\
\textit{Columbia University}\\
New York, NY 10027, USA \\
\{saa2244, my2826, bx2177\}@columbia.edu}}

\maketitle
\begin{abstract}
This paper proposes a few-shot classification framework based on Large Language Models (LLMs) to predict whether the next day will have spikes in real-time electricity prices. The approach aggregates system state information, including electricity demand, renewable generation, weather forecasts, and recent electricity prices, into a set of statistical features that are formatted as natural-language prompts and fed to an LLM along with general instructions. The model then determines the likelihood that the next day would be a spike day and reports a confidence score. Using historical data from the Texas electricity market, we demonstrate that this few-shot approach achieves performance comparable to supervised machine learning models, such as Support Vector Machines and XGBoost, and outperforms the latter two when limited historical data are available. These findings highlight the potential of LLMs as a data-efficient tool for classifying electricity price spikes in settings with scarce data.
\end{abstract}
\vspace{-0.5mm}
\vspace{-2.5mm}
\section{Introduction}\label{sec:introduction}
The short-term behavior of electricity prices has become increasingly difficult to predict as modern power systems incorporate higher levels of renewable generation and face greater variability in operating conditions. These have led to more frequent fluctuations in net load and produced highly volatile real-time price patterns~\cite{wang2017impact}. Because operators and market participants rely on accurate next-day forecasts for scheduling, hedging, and risk management, the ability to anticipate price spikes is increasingly important. In practice, many market participants rely on automated bidding and risk-management software trained on the majority of normal days. While these systems perform reliably under typical conditions, they often perform poorly during extreme price events, leading to financial losses and inefficient market responses. However, forecasting these events remains challenging due to the combined effects of operational uncertainty, renewable forecast errors, evolving demand profiles, adverse weather conditions, and persistent transmission congestion.

A wide range of statistical and machine-learning models has been applied to electricity price forecasting, with most focusing on regression-based prediction. Traditional time-series methods and modern deep-learning architectures, including Multi-Layer Perceptron (MLP)~\cite{panapakidis2016day}, hybrid neural networks~\cite{lehna2022forecasting}, and transformer-based models~\cite{alghumayjan2024energy}, aim to directly estimate future prices. Conformal prediction has also been explored to quantify predictive uncertainty~\cite{conformal}. While these methods capture general price dynamics, they struggle with rare and highly nonlinear spike events, which are overshadowed by the predominance of non-spike prices in training.

In practical operations, market participants frequently need to know whether the next day is likely to experience abnormally high prices, the ability to flag such days in advance would warrant hedging or operational adjustments. This motivates electricity spike classification, in which the goal is to determine whether prices will exceed a critical threshold. Prior studies have investigated such approaches using support vector machines (SVMs)~\cite{zhang2022short} and weighted K-nearest neighbor (WKNN) models~\cite{liu2022data}, with additional methods surveyed in~\cite{nowotarski2018recent}. Although these approaches improve sensitivity to extreme events, they typically require large labeled datasets and may degrade when spikes are infrequent or when system conditions shift over time. Such spike days also affect residential consumers, who are generally risk-averse. With the increasing adoption of virtual power plants that aggregate residential batteries for market participation~\cite{hwang2025virtual}, poor anticipation of these days can lead to inefficient bidding and financial losses.

Large language models (LLMs) can interpret structured numerical data expressed in natural language and link system signals, including demand variability, renewable output, and price trends, to price outcomes. They also perform well with limited labeled data, addressing the rarity of spike events and changing power conditions. Recently, LLMs has been used in some power system applications~\cite{lu2024large, zhou2025empower, jena2025llm, jia2025enhancing}. Motivated by this, we propose a few-shot LLM-based classifier that estimates the likelihood of next-day real-time price spikes from textual system descriptions. Our main contributions are:

\begin{figure*}[htbp]
\centerline{\includegraphics[trim = 0mm 0mm 0mm 0mm, clip, width=.85\linewidth]{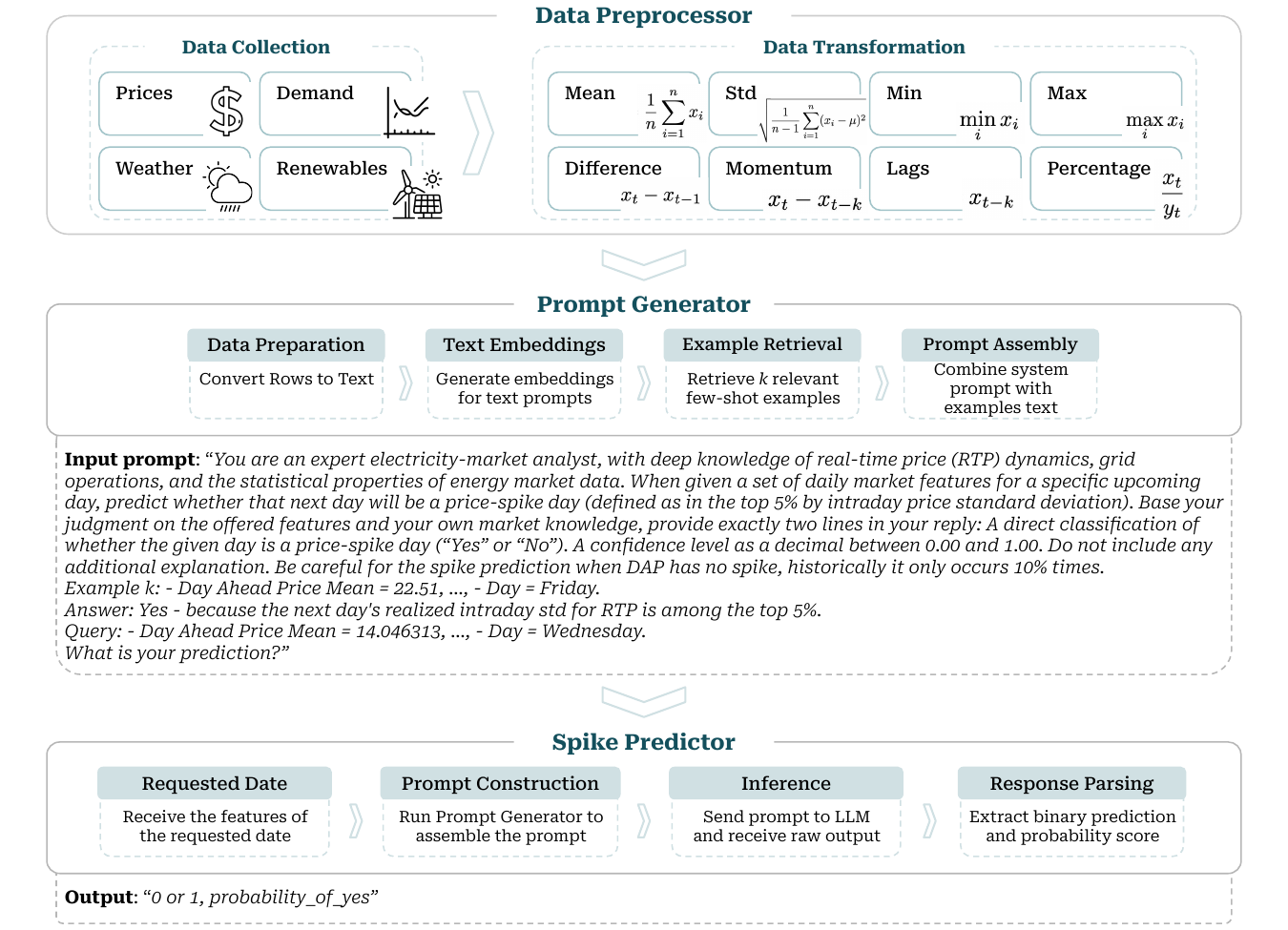}}
\vspace{-3mm}
\caption{The pipeline for the proposed approach.}
\label{fig:LLM_diagram}
\end{figure*}

\begin{itemize}
    \item We develop a few-shot classification framework that leverages LLMs to flag whether the next day will be an electricity price spike day in the real-time market, based on statistical summaries of system conditions.
    \item We introduce a simple but effective procedure that converts market and system variables, including forecasts and recent price dynamics, into textual features suitable for LLM-based inference.
    \item We evaluate the proposed framework using historical Texas market data and benchmark its performance against SVM and XGBoost, despite the latter two being trained on a much larger dataset.
    \item We demonstrate that LLM-based predictors are effective in limited-data settings, highlighting their value as a data-efficient approach to electricity price classification.
\end{itemize}
The remainder of the paper is organized as follows: Section II introduces the framework. Section III presents the case study, and Section IV concludes the paper.

\section{LLM-Based Spike Classifier Framework}\label{sec:methodology}

We present an LLM-based framework to predict whether the next day will feature real-time electricity price spikes. The framework consists of three sequential components: a \emph{data preprocessor}, which integrates and transforms electricity system wide datasets into structured learning features; a \emph{prompt generator}, which converts daily feature vectors into textual representations and retrieves relevant few-shot examples using embedding-based similarity search; and a \emph{spike predictor}, which assembles the final prompt, queries the LLM, and parses the resulting prediction. These components form a complete pipeline, spanning from raw system information to spike-day probability forecasts, without relying on model training. Figure~\ref{fig:LLM_diagram} shows a high-level diagram of the full workflow.

\subsection{Data Preprocessor}
The first component of the framework collects all the information that describes the state of the electricity system for each day. These variables are collected from several sources, including day-ahead and real-time market prices, forecasts and actual net demand, forecasted reserve levels, renewable generation forecasts, and weather data. Each dataset is aligned and transformed into a collection of engineered features intended to summarize system conditions. These include daily statistics such as means and standard deviations, lagged values that incorporate information from the previous day, and multi-day summaries that capture recent trends. Weather-related measures, renewable penetration, and calendar indicators are also included to account for environmental and temporal patterns. We define a spike day as any day whose real-time intraday price standard deviation exceeds the $95\%$ threshold, where
\[
\text{Spike}_t =
\begin{cases}
1, & \text{if } \sigma^{\text{RTP}}_{\text{intraday},t} > 
\mathrm{Quantile}_{0.95}\!\left(\sigma^{\text{RTP}}_{\text{intraday}}\right), \\[6pt]
0, & \text{otherwise},
\end{cases}
\]
Table~\ref{tab:feature_table} provides an overview of all engineered features, categorized by their underlying data source. This feature set forms the daily feature vector that is later translated into text for use in the prompt generation stage.

\begin{table}[htbp]
\centering
\caption{Engineered features by Data Category}
\footnotesize
\begin{tabular}{c l}
\toprule
\textbf{Category} & \textbf{Feature} \\
\midrule

\multirow{7}{*}{\centering Day-Ahead Price}
 & \texttt{dap\_mean} \\
 & \texttt{dap\_std} \\
 & \texttt{dap\_max} \\
 & \texttt{dap\_min} \\
 & \texttt{dap\_momentum} \\
 & \texttt{dap\_mean\_3d} \\
 & \texttt{dap\_std\_3d} \\
\midrule

\multirow{2}{*}{\centering Real-Time Price}
 & \texttt{rtp\_mean\_lag1} \\
 & \texttt{rtp\_std\_lag1} \\
\midrule

\multirow{5}{*}{\centering Net Demand (Forecast)}
 & \texttt{net\_demand\_forecast\_mean} \\
 & \texttt{net\_demand\_forecast\_min} \\
 & \texttt{net\_demand\_forecast\_max} \\
 & \texttt{net\_demand\_forecast\_std} \\
 & \texttt{net\_demand\_diff} \\
\midrule

\multirow{2}{*}{\centering Net Demand (Actual)}
 & \texttt{net\_demand\_actual\_mean\_lag1} \\
 & \texttt{net\_demand\_actual\_std\_lag1} \\
\midrule

\multirow{4}{*}{\centering Forecasted Reserves}
 & \texttt{reserve\_mean} \\
 & \texttt{reserve\_std} \\
 & \texttt{reserve\_max} \\
 & \texttt{reserve\_min} \\
\midrule

\multirow{5}{*}{\centering Weather}
 & \texttt{temp\_max} \\
 & \texttt{temp\_min} \\
 & \texttt{temp\_diff} \\
 & \texttt{wind\_speed\_max} \\
 & \texttt{wind\_speed\_min} \\
\midrule

\multirow{1}{*}{\centering Renewables}
 & \texttt{pct\_renewable\_forecast} \\
\midrule

\multirow{2}{*}{\centering Calendar}
 & \texttt{day} \\
 & \texttt{month} \\
 \midrule

\multirow{2}{*}{\centering Labels}
 & \texttt{is\_dap\_spike} \\
 & \texttt{is\_spike} \\
\bottomrule
\end{tabular}
\label{tab:feature_table}
\end{table}

\subsection{Prompt Generator}
The second component of the framework converts each day’s feature vector into natural-language descriptions suitable for use in large language models. For each row, the feature values are transformed into a textual prompt. These text prompts are then mapped into numerical embedding using an embedding model. To efficiently locate past days that resemble the query day, we use FAISS~\cite{douze2024faiss}, a library designed for fast similarity search in large collections of vectors.

FAISS builds an index over the embeddings that enables quick nearest-neighbor queries, even when the dataset is large. Using this index, the prompt generator identifies a set of candidate examples that are most similar to the day being predicted. When predicting for a query date, the prompt generator retrieves the most similar historical spike day and then selects the closest days to form a candidate pool. From these candidates, a set of \(k\) examples is selected using maximal marginal relevance (MMR)~\cite{carbonell1998use}, which helps choose examples that are relevant to the query and not redundant with one another. The final prompt provided to the LLM consists of the system instruction, the selected examples with their labels, and the text description of the query day. The full procedure is summarized in Algorithm~\ref{alg:promptgen}.

\begin{algorithm2e}
\caption{Prompt Generator}\label{alg:promptgen}
\KwData{Input dataframe $\mathbf{D}$}
\textbf{Require}: System prompt $S$, number of examples $k$, MMR parameter $\lambda$, embedding model $\mathcal{E}$\\
\textbf{Data Preparation}\\
\ForEach{row $r$ in $\mathbf{D}$}{Convert feature values of $r$ into a textual description $t_r$;}
\textbf{Text Embeddings}\\
\ForEach{text prompt $t_r$}{Compute embedding $e_r$ using model $\mathcal{E}$;}
Store embeddings $\{e_r\}$;\\
\textbf{Example Retrieval}\\
Normalize embeddings and construct FAISS index using $\{e_r\}$;\\
Convert query row to text prompt $t_q$ and compute embedding $e_q$;\\
Identify valid prior rows $r < q$;\\
Select the most similar prior spike example;\\
Retrieve additional nearest-neighbor candidates from the FAISS index;\\
\textbf{Apply Maximal Marginal Relevance (MMR)}:\\
Initialize selected set $\mathcal{S}$;\\
\While{$|\mathcal{S}| < k$}{
    Compute MMR score for each candidate using query similarity and redundancy penalty $\lambda$;\\
    Add the highest-scoring candidate to $\mathcal{S}$;
}
\textbf{Prompt Assembly}\\
\ForEach{example $e_i \in \mathcal{S}$}{Format example text and associated label;\\}
Append the query description;\\
Concatenate system prompt $S$, retrieved examples, and query description;\\
\Return Final prompt ready for LLM inference.
\end{algorithm2e}

\subsection{Spike Predictor}
The final component executes LLM-based inference. Given a requested date, the spike predictor obtains the corresponding feature vector, calls the prompt generator to construct the full textual prompt, and sends this prompt to the LLM. The LLM returns a structured two-line response: a binary classification (\texttt{Yes}/\texttt{No}) indicating whether the next day is expected to be a spike day, and a confidence score between 0 and 1. The predictor parses this output to produce the final prediction and associated probability.

\section{Case Study}

\begin{figure*}[htbp]
\centerline{\includegraphics[width=0.9\linewidth]{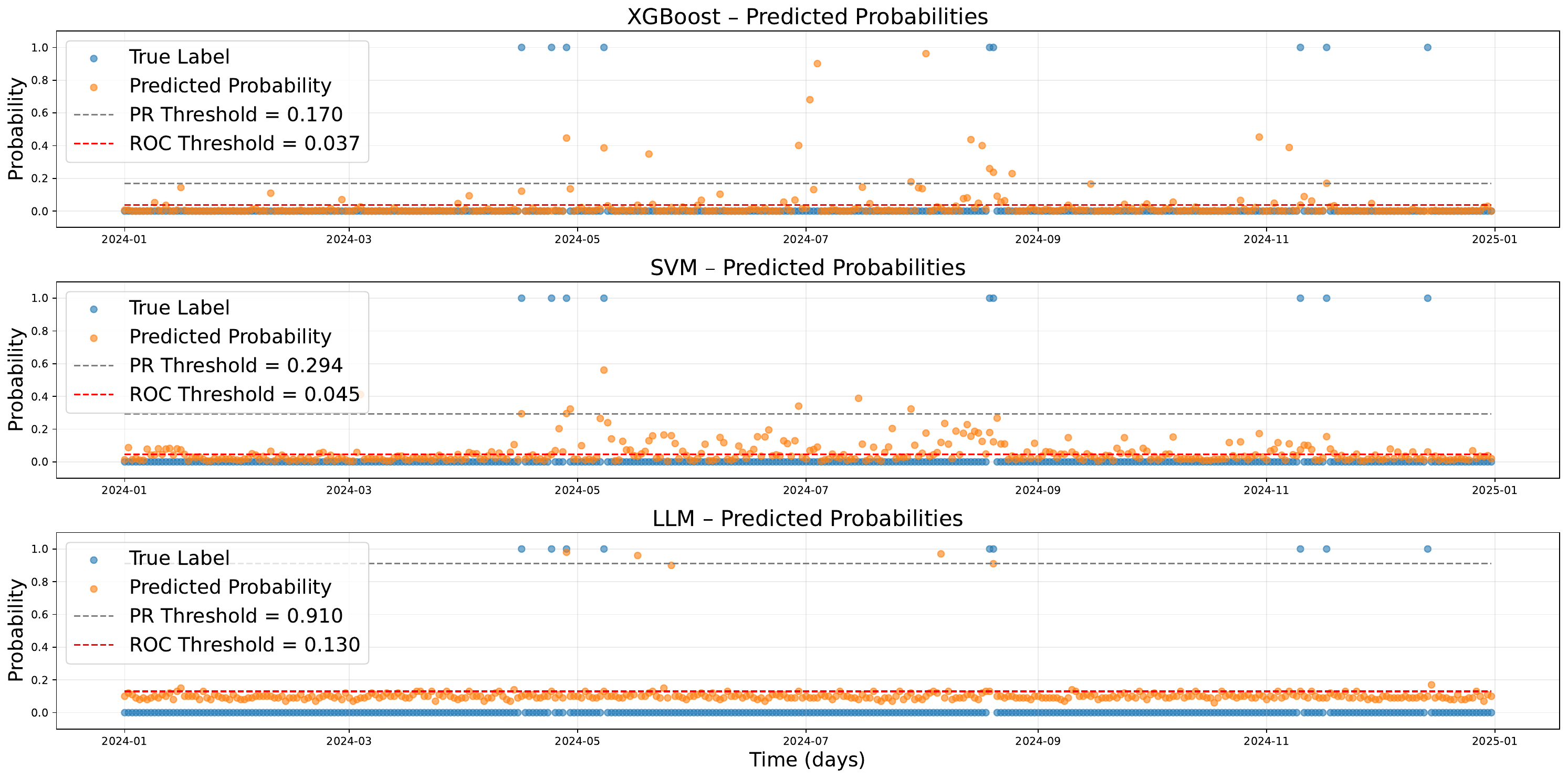}}
\caption{Prediction probabilities for XGBoost, SVM, and LLM models, with PR and ROC thresholds shown for comparison.}
\label{fig:probabilities}
\end{figure*}

This section evaluates the proposed LLM-based spike classification framework using historical data from the Texas electricity market. We describe the experimental setup, the classification metrics used for performance assessment, and the results obtained across multiple thresholding strategies and data availability scenarios.

\subsection{Experiment Setup}

We utilize the Electric Reliability Council of Texas (ERCOT) system-wide data covering the period from 2021 to 2024. The years 2021-2023 are used for training and 2024 is used for testing. All models produce a one-day-ahead prediction of whether the next day will be a spike day. The engineered features described in Section~\ref{sec:methodology} serve as inputs to the prompt generator. For the LLM-based classifier, we use the \texttt{gpt-4.1} model~\cite{openai_gpt4_1}. Benchmark models include SVM~\cite{cristianini2008support} and XGBoost~\cite{chen2016xgboost} classifiers, both trained on the same feature set. All computations were carried out on a high-performance computing cluster with Intel Xeon Platinum 8640Y 2Ghz CPUs, 512GB memory, and 2xNVIDIA A40 GPUs.

\subsection{Classification Metrics}

To evaluate the performance of the spike classifiers, we rely on standard measures derived from the confusion matrix. Let TP, FP, FN, and TN denote the numbers of true positives, false positives, false negatives, and true negatives:
\begin{align}
\text{Precision} &= {\mathrm{TP}}/{(\mathrm{TP} + \mathrm{FP})} \\
\text{Recall} &= {\mathrm{TP}}/{(\mathrm{TP} + \mathrm{FN})}\\
\text{Accuracy} &= {\mathrm{TP} + \mathrm{TN}}/{(\mathrm{TP} + \mathrm{FP} + \mathrm{FN} + \mathrm{TN})} \\
\text{F1-score} &= 2 \cdot {\text{Precision} \cdot \text{Recall}}/{(\text{Precision} + \text{Recall})}
\end{align}

Precision reflects the fraction of predicted spikes that are correct, while recall measures the fraction of actual spike days correctly identified. Accuracy provides the overall proportion of correct predictions, and the F1-score summarizes the balance between precision and recall.

In addition to these point metrics, we examine the receiver operating characteristic (ROC) curve and the precision–recall (PR) curve. The ROC curve plots the true positive rate against the false positive rate as the classification threshold varies, highlighting the classifier’s ability to separate spike and non-spike days across all thresholds.

The PR curve plots precision against recall across thresholds. This curve informs settings where rare positive events occur. Because spike days constitute a small fraction of the dataset, precision–recall analysis provides a better understanding of the classifier behavior. We examine both ROC and PR curves and derive threshold values from each: the ROC-optimal threshold balances sensitivity and false-alarm rate, and the PR-optimal threshold maximizes the F1-score.

\subsection{Results}

Table~\ref{tab:main_table} summarizes the classification outcomes under the three thresholding settings. When using the full training set, the LLM performs competitively across several metrics despite not being trained directly on historical labels. In particular, the LLM often achieves the highest or second-highest among all metrics under multiple thresholding scenarios. These results indicate that the LLM can effectively leverage prompt-generated context to detect spike events, comparable to supervised learning models that rely on extensive training data.

\begin{table}[htbp]
\caption{Model Performance Across Threshold Types}
\scriptsize
\begin{center}
\begin{tabular}{c c c c c c c c c}
\toprule
Model & TP & FP & FN & TN & Prec. & Rec. & Acc. & F1 \\
\midrule
\multicolumn{9}{c}{\textbf{Fixed (0.5)}} \\
\multicolumn{9}{c}{ Thresholds: XGBoost = 0.50; SVM = 0.50; LLM = 0.50;} \\
\cmidrule(lr){1-9}
XGBoost & 0.0 & 3.0 & 9.0 & 354.0 & 0.0 & 0.0 & 96.72 & 0.0 \\
SVM & 1.0 & 0.0 & 8.0 & 357.0 & \textbf{100.0} & \underline{11.11} & \textbf{97.81} & \underline{20.0} \\
LLM & 2.0 & 3.0 & 7.0 & 354.0 & \underline{40.0} & \textbf{22.22} & \underline{97.27} & \textbf{28.57} \\
\midrule
\multicolumn{9}{c}{\textbf{ROC-optimal}} \\
\multicolumn{9}{c}{ Thresholds: XGBoost = 0.04; SVM = 0.04; LLM = 0.13;} \\
\cmidrule(lr){1-9}
XGBoost & 6.0 & 43.0 & 3.0 & 314.0 & \underline{12.24} & \underline{66.67} & \underline{87.43} & \underline{20.69} \\
SVM & 9.0 & 140.0 & 0.0 & 217.0 & 6.04 & \textbf{100.0} & 61.75 & 11.39 \\
LLM & 6.0 & 37.0 & 3.0 & 320.0 & \textbf{13.95} & \underline{66.67} & \textbf{89.07} & \textbf{23.08} \\
\midrule
\multicolumn{9}{c}{\textbf{PR-optimal}} \\
\multicolumn{9}{c}{ Thresholds: XGBoost = 0.17; SVM = 0.29; LLM = 0.91;} \\
\cmidrule(lr){1-9}
XGBoost & 4.0 & 11.0 & 5.0 & 346.0 & 26.67 & \textbf{44.44} & 95.63 & \underline{33.33} \\
SVM & 3.0 & 5.0 & 6.0 & 352.0 & \underline{37.5} & \underline{33.33} & \underline{96.99} & \textbf{35.29} \\
LLM & 2.0 & 2.0 & 7.0 & 355.0 & \textbf{50.0} & 22.22 & \textbf{97.54} & 30.77 \\
\bottomrule
\end{tabular}
\end{center}
\label{tab:main_table}
\end{table}
Figure~\ref{fig:probabilities} plots the predicted spike probabilities for each model across the test period, along with the PR-optimal and ROC-optimal thresholds. The LLM produces smoother probabilities and tends to concentrate high-confidence predictions around periods of spikes, while the SVM and XGBoost models exhibit more fluctuation that reflect less confident on predictions. This explains why the LLM model has higher thresholds than the other two models. 

Finally, Table~\ref{tab:limited} reports performance in a limited-data setting, where only two months of historical data are used for training, instead of the full three years. In this scenario, the LLM considerably outperforms the supervised baselines across most metrics. Whereas  SVM and XGBoost suffer substantial degradation when training data are scarce, the LLM remains stable. This demonstrates a key advantage of the proposed approach when labeled training data are highly limited.

\begin{table}[t]
\caption{Model Performance Across Threshold Types - Limited Data}
\scriptsize
\begin{center}
\begin{tabular}{c c c c c c c c c}
\toprule
Model & TP & FP & FN & TN & Prec. & Rec. & Acc. & F1 \\
\midrule
\multicolumn{9}{c}{\textbf{Fixed (0.5)}} \\
\multicolumn{9}{c}{ Thresholds: XGBoost = 0.50; SVM = 0.50; LLM = 0.50;} \\
\cmidrule(lr){1-9}
XGBoost & 1.0 & 74.0 & 8.0 & 283.0 & 1.33 & \underline{11.11} & 77.6 & 2.38 \\
SVM & 0.0 & 0.0 & 9.0 & 357.0 & 0.0 & 0.0 & \textbf{97.54} & 0.0 \\
LLM & 3.0 & 7.0 & 6.0 & 350.0 & \textbf{30.0} & \textbf{33.33} & \underline{96.45} & \textbf{31.58} \\
\midrule
\multicolumn{9}{c}{\textbf{ROC-optimal}} \\
\multicolumn{9}{c}{ Thresholds: XGBoost = 1.00; SVM = 0.03; LLM = 0.93;} \\
\cmidrule(lr){1-9}
XGBoost & 0.0 & 0.0 & 9.0 & 357.0 & 0.0 & 0.0 & \textbf{97.54} & 0.0 \\
SVM & 2.0 & 75.0 & 7.0 & 282.0 & 2.6 & \underline{22.22} & 77.6 & 4.65 \\
LLM & 3.0 & 5.0 & 6.0 & 352.0 & \textbf{37.5} & \textbf{33.33} & \underline{96.99} & \textbf{35.29} \\
\midrule
\multicolumn{9}{c}{\textbf{PR-optimal}} \\
\multicolumn{9}{c}{ Thresholds: XGBoost = 0.03; SVM = 0.03; LLM = 0.93;} \\
\cmidrule(lr){1-9}
XGBoost & 9.0 & 357.0 & 0.0 & 0.0 & 2.46 & \textbf{100.0} & 2.46 & 4.8 \\
SVM & 2.0 & 75.0 & 7.0 & 282.0 & 2.6 & 22.22 & \underline{77.6} & 4.65 \\
LLM & 3.0 & 5.0 & 6.0 & 352.0 & \textbf{37.5} & \underline{33.33} & \textbf{96.99} & \textbf{35.29} \\
\bottomrule
\end{tabular}
\end{center}
\label{tab:limited}
\end{table}
\vspace{-1mm}

\section{Conclusion}
This paper introduces an LLM-based framework for predicting whether the next day will be a day of spikes in real-time electricity prices using a few-shot classification approach. The method integrates system-level data, engineered features, and a structured prompt design to enable large language models to perform spike classification without direct model training. Through a combination of textual feature representation, similarity-based example retrieval, and LLM inference, the proposed approach achieves performance comparable to established machine learning models despite having no supervised training component. Our results demonstrate that, while traditional models such as XGBoost and SVM perform well when large training datasets are available, their performance deteriorates substantially under limited-data conditions, during which the LLM-based classifier remains competitive, highlighting its value as a data-efficient classification tool. For the next steps, several extensions are possible, such as expanding the prompt format to include multi-day contextual information or fine-tuning domain-specific LLMs, which may further enhance spike prediction accuracy.

\bibliographystyle{IEEEtran}
\bibliography{references}

\end{document}